\documentclass[conference]{IEEEtran}
\IEEEoverridecommandlockouts
\usepackage{cite}
\usepackage{amsmath,amssymb,amsfonts}
\usepackage{algorithmic}
\usepackage{graphicx}
\usepackage{textcomp}
\usepackage{xcolor,colortbl}

\usepackage[thinlines]{easytable}
\usepackage{stackengine}
\usepackage{array}

\usepackage{ragged2e}
\usepackage{graphicx}
\usepackage[export]{adjustbox}

\usepackage{blindtext}

\usepackage{adjustbox}
\usepackage{multirow}

\usepackage{breakcites}
\usepackage{cite}
\usepackage{bm}
\usepackage{colortbl}

\usepackage{mathtools}

\usepackage[center]{caption}

\usepackage[edges]{forest}

\usepackage{float}

\usepackage{inconsolata}
 \usepackage{url}

\usepackage{graphicx}
\usepackage[export]{adjustbox}

\definecolor{Gray}{gray}{0.50}
\definecolor{Gray2}{gray}{0.70}

\usetikzlibrary{fit}

\definecolor{burgundy}{rgb}{0.5, 0.0, 0.13}
\definecolor{brownweb}{rgb}{0.65, 0.16, 0.16}
\definecolor{chromeyellow}{rgb}{1.0, 0.65, 0.0}
\definecolor{babypink}{rgb}{0.96, 0.76, 0.76}
\definecolor{melon}{rgb}{0.99, 0.74, 0.71}
\definecolor{moccasin}{rgb}{0.98, 0.92, 0.84}

\def\BibTeX{{\rm B\kern-.05em{\sc i\kern-.025em b}\kern-.08em
    T\kern-.1667em\lower.7ex\hbox{E}\kern-.125emX}}
\begin{document}

\title{Damaged Fingerprint Recognition by Convolutional Long Short-Term Memory Networks for  Forensic Purposes
}

\author{\IEEEauthorblockN{Jaouhar Fattahi and Mohamed Mejri}

\IEEEauthorblockA{{Department of Computer Science and Software Engineering.}\\
{Laval University, }Quebec city, Canada.\\
{2325 rue de l'Universit\'e, Quebec city, QC, CA, G1V 0A6.}\\
Emails: jaouhar.fattahi.1@ulaval.ca; mohamed.mejri@ift.ulaval.ca}

}

\maketitle

\begin{abstract}

Fingerprint recognition is often a game-changing step in establishing evidence against criminals. However, we are increasingly finding that criminals deliberately alter their fingerprints in a variety of ways to make it difficult for technicians and automatic sensors to recognize their fingerprints, making it tedious for investigators to establish strong evidence against them in a forensic procedure. In this sense, deep learning comes out as a prime candidate to assist in the recognition of damaged fingerprints. In particular, convolution algorithms. In this paper, we focus on the recognition of damaged fingerprints by Convolutional Long Short-Term Memory networks. We present the architecture of our model and demonstrate its performance which exceeds 95\% accuracy, 99\% precision, and approaches 95\% recall and 99\% AUC.\\

\end{abstract}


\textit{This paper  was  accepted, on December 5, 2020, for publication and oral presentation at the 2021 IEEE 5th International Conference on Cryptography, Security and Privacy (CSP 2021) to be held in Zhuhai, China during January 8-10, 2021 and hosted by Beijing Normal University (Zhuhai).}

\section*{Notice}

\textbf{\copyright 2021 IEEE. \textit{Personal use of this material is permitted. Permission from IEEE must be obtained for all other uses, in any current or future media, including reprinting/republishing this material for advertising or promotional purposes, creating new collective works, for resale or redistribution to servers or lists, or reuse of any copyrighted component of this work in other works.}}

\section{Introduction}

Fingerprints are traces that we leave behind every time we touch an object. The patterns created by the ridges and folds of the the fingers' skin are different for each person, and can therefore be used to identify someone in a virtually unique and permanent way, just like other biometric data such as DNA, iris, face, voice, etc. Fingerprints can serve an important role in the identification of victims of both natural and man-made disasters \cite{BECUE2020}, such as earthquakes, floods, civil strife, and terrorist attacks. This proof of identity is critical not only for the investigators \cite{DBLPHerrmannFF14, DBLPMarcialisRCD10}, but also for the families of the victims concerned. When a fingerprint is taken from a crime scene, it is compared to fingerprints stored in police databases to prove whether or not a suspect is involved in multiple actions. On the other hand, fingerprint identification is the cornerstone of national security and is the most practical biometric means for the identification of criminals, drug dealers, human traffickers, immigration law violators, terrorists, who can infiltrate the territory under false identities through the ports of entry of a given country. Similarly, fingerprints are increasingly being used as a means of choice to confirm a person's identity in access and payment transactions \cite{DBLPMoalosiHP19}. This method is indeed much faster, easier and more secure than the traditional use of passwords or PIN codes, which have the disadvantages of being difficult to remember and easy to steal. For their part, criminals' means have constantly and rapidly evolved over time, allowing them to evade the traditional means of fingerprint recognition installed at ports of entry and police checkpoints \cite{DBLPYoonFJ12, Josphineleela112233}. Among these means, criminals often choose to deliberately damage the skin of their fingers to obfuscate their fingerprints so that they appear different, for recognition algorithms, from those stored in databases. In this complex context, artificial intelligence becomes the tool of choice for identifying damaged fingerprints and their owners. In this paper, we address the problem of recognizing damaged fingerprints using Convolutional Long Short-Term Memory networks (Convolutional LSTMs) \cite{doi10116201199}. In fact, an image of a fingerprint can be seen as a set of patterns (or features) that are captured by convolutional operations. It is also seen as a sequence of entities (pixels, rows, columns, features). This notion of sequence is generally well suited to processing by specialized recurrent neural networks such as Long Short-Term Memory (LSTM) networks. This makes the subject a good candidate to be addressed by Convolutional LSTMs.

\section{Damaged fingerprints} \label{sec211}

Damage to fingerprints is often deliberate and of several sources and types\cite{DBLPFengJR10}. It is mainly caused by the use of strong chemicals or by physical actions and lesions applied to the surface of fingers. This damage results in severely breaking the link between a person's fingerprints and their identity. In this regard, it is worth emphasizing that the problem of damaged fingerprints differs substantially from the problem of falsified fingerprints, which involves fake fingerprints made of a material that sticks to the palms of the fingers to divert recognition software. In this paper, we will consider the following alterations \cite{DBLPSOKOTO} that can muddle most of fingerprint verification software.

\subsection{Obliteration}

\begin{figure}[htbp]
\centering
\begin{minipage}{3cm}
\includegraphics[width=2.5cm,height=2.5cm]{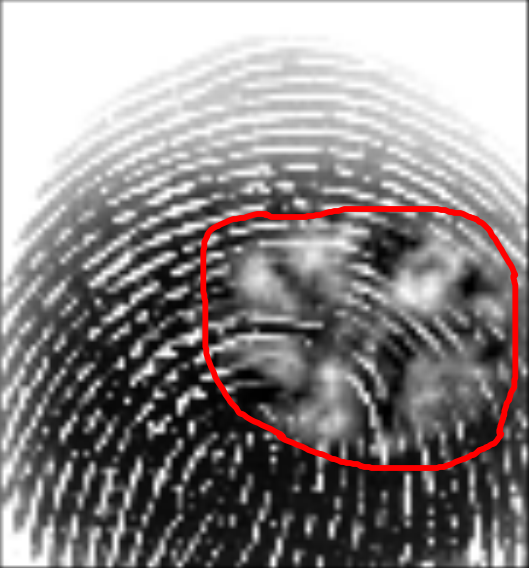}
\end{minipage}
\caption{Obliterated fingerprint  \cite{DBLPSOKOTO}}
\label{figOBL}
\end{figure}

Fingerprints on fingertips can be obliterated by various means such as burning, abrasion, application of strong chemicals and skin transplantation which cause an alteration of the morphology of the surface of the fingertips coming from a strong and damaging action on the skin cells. Fig. \ref{figOBL} shows an example of a fingerprint being obliterated using a chemical substance.

\subsection{Z-cut}

\begin{figure}[htbp]
\centering
\begin{minipage}{3cm}
\includegraphics[width=2.5cm,height=2.5cm]{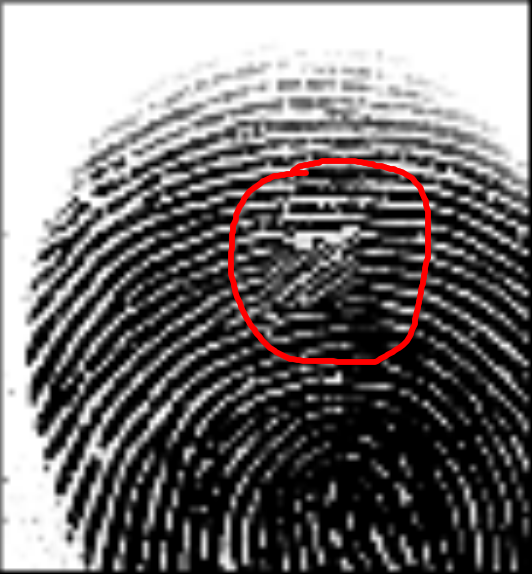}
\end{minipage}
\caption{Z-cut fingerprint  \cite{DBLPSOKOTO}}
\label{figZcut}
\end{figure}

A Z-cut is a lesion in the shape of the letter Z brought to the skin of the fingertip. This type of modification of fingerprints is increasingly observed in border control applications by suspicious travellers and illegal asylum seekers to bypass identification systems which are generally vulnerable to such manoeuvres, some wanting to evade existing regulations, others wanting access to the priority asylum procedure. Fig. \ref{figZcut} shows a fingerprint with a Z-cut.

\subsection{Central rotation }

\begin{figure}[htbp]
\centering
\begin{minipage}{3cm}
\includegraphics[width=2.5cm,height=2.5cm]{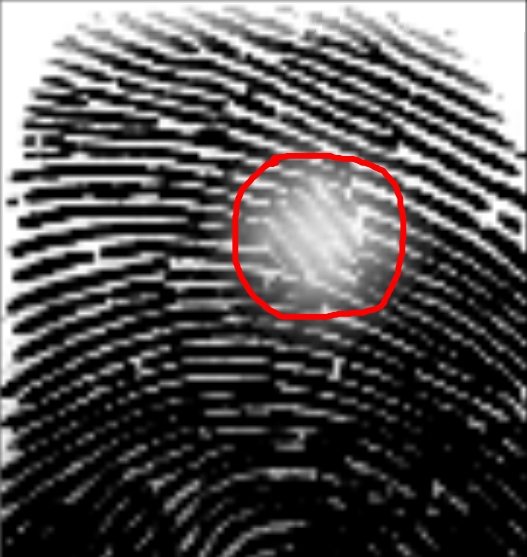}
\end{minipage}
\caption{Central rotated fingerprint  \cite{DBLPSOKOTO}}
\label{figCR}
\end{figure}

This alteration consists of removing a slice of the skin from a finger and planting it in a different position in the finger's tip in a rotated way. The means to do this is not too complex and the resulting altered fingerprints are very hard to compare with the original fingerprints by automated matching systems. Fig. \ref{figCR} shows a fingerprint with a central rotation.

\section{Deep Learning Algorithms} \label{sec5}

\subsection{Long Short-Term Memory Networks}

\begin{figure}[!htb]
\centering
\includegraphics[scale=0.3, frame]{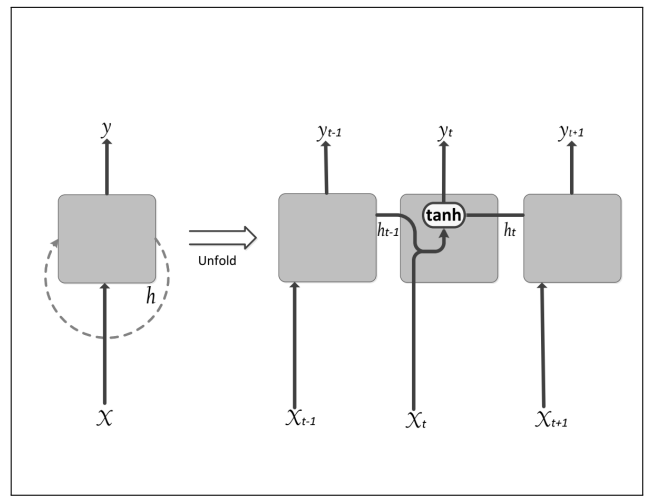}

\caption{RNN process}
\label{RNN}
\end{figure}

Long Short-Term Memory  (LSTM)  networks \cite{hochreiter1997long, SHERSTINSKY2020132306} are basically Recurring Neural Networks (RNNs). RNNs take into account not only the current input but also the previous output (see Fig \ref{RNN}). That is to say, the decision made at the step $t-1$ affects the decision that will be made at the next step $t$. In other words, a standard RNN (aka vanilla RNN) is a neural network with memory. Adding memory to RNNs makes them powerful in processing sequences such as images, sound, videos, texts, etc. The sequential information is maintained in an internal memory (hidden state)  that captures long-term dependencies between events that may be distant in time.  Although RNNs manage sequences efficiently, they suffer from a number of issues \cite{Bengio}. The first is the exploding problem. This occurs when large  gradient errors build up  as the training progresses, which leads to very large updates in the network weights and resulting in an unstable network. The second problem is the vanishing gradient. This occurs when the gradient values become extremely low, causing the model to stop learning. Moreover, RNNs are known to be sensitive to the way the model is initialized. This can lead to model overfitting and thus strange results may occur. 

LSTMs, on the other hand, were designed to overcome the drawbacks of RNNs. The key element of an LSTM is its cell state, which acts as the memory of the network that holds the ad rem  information during sequence processing. In basic RNNs, the repetition module consists of a single hyperbolic tangent layer \texttt{tanh} \cite{DBLPActivationFunctions03378, DBLPMarraIM07}. In LSTMs, the repetition module is much more complex and made up of three elements referred to as \textit{gates}. The first gate is called forget gate, which consists of a sigmoid \cite{DBLPActivationFunctions03378} neural network layer.

\begin{figure}[!htb]
\centering
\includegraphics[scale=0.3, frame]{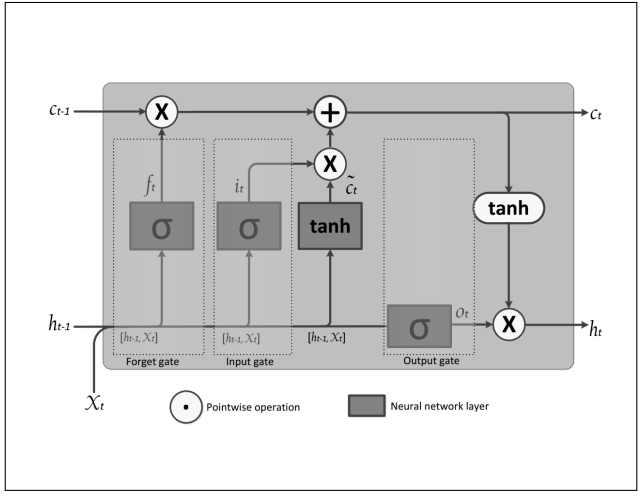}
\caption{LSTM (cell)}
\label{LSTM}
\end{figure}

At a step $t$, this layer looks at the previous hidden output $h_{t-1}$ as well as the current input $x_t$. These two values are first concatenated, then multiplied by weight matrices, then added to a bias value. After that, a sigmoid function is applied. This is given by  Equation \ref{eq100}.

\begin{equation} f_t  =  \sigma(W_f . [h_{t-1},x_t] + b_f)  \label{eq100} \end{equation}

The sigmoid function generates a vector with zeros and ones.  Essentially, its role is to decide which values the cell state should retain (for ones) and which values should simply discard (for zeros). This vector is then multiplied (pointwise product) by the previous cell state $c_{t-1}$.

The second gate is referred to as the input gate. Its purpose is to decide which new information needs to be added to the cell state. This is achieved through two layers. The first one is a sigmoid layer that issues a vector of zeros and one that determine what should be modified and what should remain, as given by Equation \ref{eq101}.
 \begin{equation}  i_t=\sigma(W_i . [h_{t-1},x_t] + b_i)  \label{eq101}\end{equation}

The second layer is a \texttt{tanh} layer that generates a vector of a candidate vector $\tilde{c}$ to add to the  cell state, as given by Equation \ref{eq102}.

\begin{equation}  \tilde{c}_t = \texttt{tanh}(W_c . [h_{t-1},x_t] + b_c)   \label{eq102}\end{equation}

It is worth noticing that the \texttt{tanh} function \cite{DBLPActivationFunctions03378} is preferred to other activation functions because it emits both positive and negative values, so that the cell state can increase or decrease.

Then, the cell state is updated considering the two previous gates as given by Equation  \ref{eq103}.

\begin{equation}  c_t = (f_t \otimes c_{t-1}) \oplus (i_t \otimes \tilde{c}_t)  \label{eq103}   \end{equation}

The third gate is referred to as the output gate. This gate determines what the cell will output. The output hinges upon a filtered version of the cell state. To do so, a sigmoid layer, comparable to the sigmoid layers in the previous gates, will first decide which parts of the cell state will be kept by emitting  a vector of zeros and ones. Then, a \texttt{tanh} function is applied to the updated cell state. At last, a pointwise product is performed on the sigmoid output vector and the filtered cell state resulting in the output state $h_t$. This is given by  Equation  \ref{eq104} and Equation \ref{eq105}. 

\begin{equation} o_t = \sigma(W_o . [h_{t-1},x_t] + b_o)      \label{eq104}    \end{equation}
\begin{equation}  h_t = o_t \otimes  \texttt{tanh} (c_t)      \label{eq105}  \end{equation}

\subsection{Convolutional Neural Networks}

A standard  Convolutional Neural Network (CNN) \cite{CNN2999257, li2020survey} is a powerful Deep Learning model when dealing with images. It is made up of four layers where the output of one layer is the input of the next:

\begin{enumerate}

\item The convolution layer (CONV): this is the most important component of the model in that it identifies a set of features in the data received using filters. The idea is to slip a small window (filter) representing the feature along the data matrix and calculate the convolution product. Pairs (matrix, filter) are then obtained (feature map). The higher the convolution product is, the more likely the feature position in the image is;

\item The pooling layer (POOL): it is interposed between two convolution layers. It receives several input feature maps, and applies to each of them a pooling operation which consists in shrinking the size of the input matrices, while preserving their important features. To achieve this, the input matrix is divided into regular cells, and then the maximum value is kept within each cell. In practice, small square cells are used to preclude losing too much information. The same number of feature maps at the output is conserved at the input, but the matrices are much smaller. The pooling layer cuts down the number of parameters and accelerates calculations. This improves network efficiency and helps to reduce overfitting; 

\item The Rectified Linear Unit (ReLU) correction layer: this layer substitutes all negative values with zeros using the ReLU activation function (Equation  \ref{eq200}).

\begin{equation}  \texttt{ReLU} (x) = \texttt{max}(0,x)  \label{eq200} \end{equation}  

This results in a neural network much faster without significantly decreasing its accuracy \cite{DBLPPretoriusBD19,DBLPXuWCL15}. This also enhances non-linearity inside the network. The whole process involving these three steps can be repeated many times.

\item The fully-connected layer (FC): This layer is similar to any ordinary layer in any neural network. It receives an input vector (usually flattened) and applies  an activation function, to finally vote for the classes.

\end{enumerate}

\subsection{Convolutional LSTM}

A Convolutional LSTM model is roughly a CNN connected to an LSTM \cite{DBLPCourtneyS19}. Hence, the image passes through the convolutions layers and results in a vector flattened to a 1D array with the obtained features forming the input of the LSTM. This being said, in some implementations,  internal matrix multiplications are exchanged with convolution operations where the data flows through the cells in two or three dimensions.

\section{Experiment setup} \label{sec6}

\subsection{Dataset}

First of all, it is worth mentioning that there is an an observable dearth of public datasets containing damaged fingerprint images, which has hampered research in this regard. In our study, we use the Sokoto Coventry Fingerprint (SOCOFing) dataset  \cite{DBLPSOKOTO}. This dataset is made up of 6000 fingerprint images taken from 600 male and female subjects. Fingerprints are synthetically altered with three different types of alteration: obliteration, Z-cut, and central rotation.  

\subsection{Data preprocessing}

Fingerprint images are read with a single grayscale channel. Then, the resulting image float arrays are normalized. The third of the dataset is reserved for validation and the rest of the dataset is used for training. We use  the \textit{stratify} strategy when splitting data so that the validation dataset distribution, as well as the training dataset distribution, remain the same as the original one.

\subsection{Model architecture}

The first layer is a Convolutional LSTM layer with 64 filters and a ReLU activation function\cite{DBLPActivationFunctions03378}. The second layer is a Dropout layer to ignore some number of layer outputs to prevent overfitting. The third layer is a Dense layer with 100 nodes and a ReLU activation function. The last layer is a Dense layer with the number of subjects nodes and a Softmax activation function  \cite{DBLPActivationFunctions03378}. Adam optimizer is selected \cite{kingma20171111adam}.

\section{Results and discussion}\label{sec7}

To evaluate how well our model is performing, we use the following metrics:

\begin{center}
\begin{equation} \text{Accuracy} = \displaystyle \frac{\text{TP+TN}}{\text{TP + TN + FP + FN}} \end{equation} 
\end{center}

\begin{center}
\begin{equation} \text{Precision} =\displaystyle \frac{\text{TP}}{\text{TP + FP}} \end{equation} 
\end{center}

\begin{figure}[htbp]
\centering
\begin{minipage}{9cm}
\includegraphics[width=9cm,height=6cm]{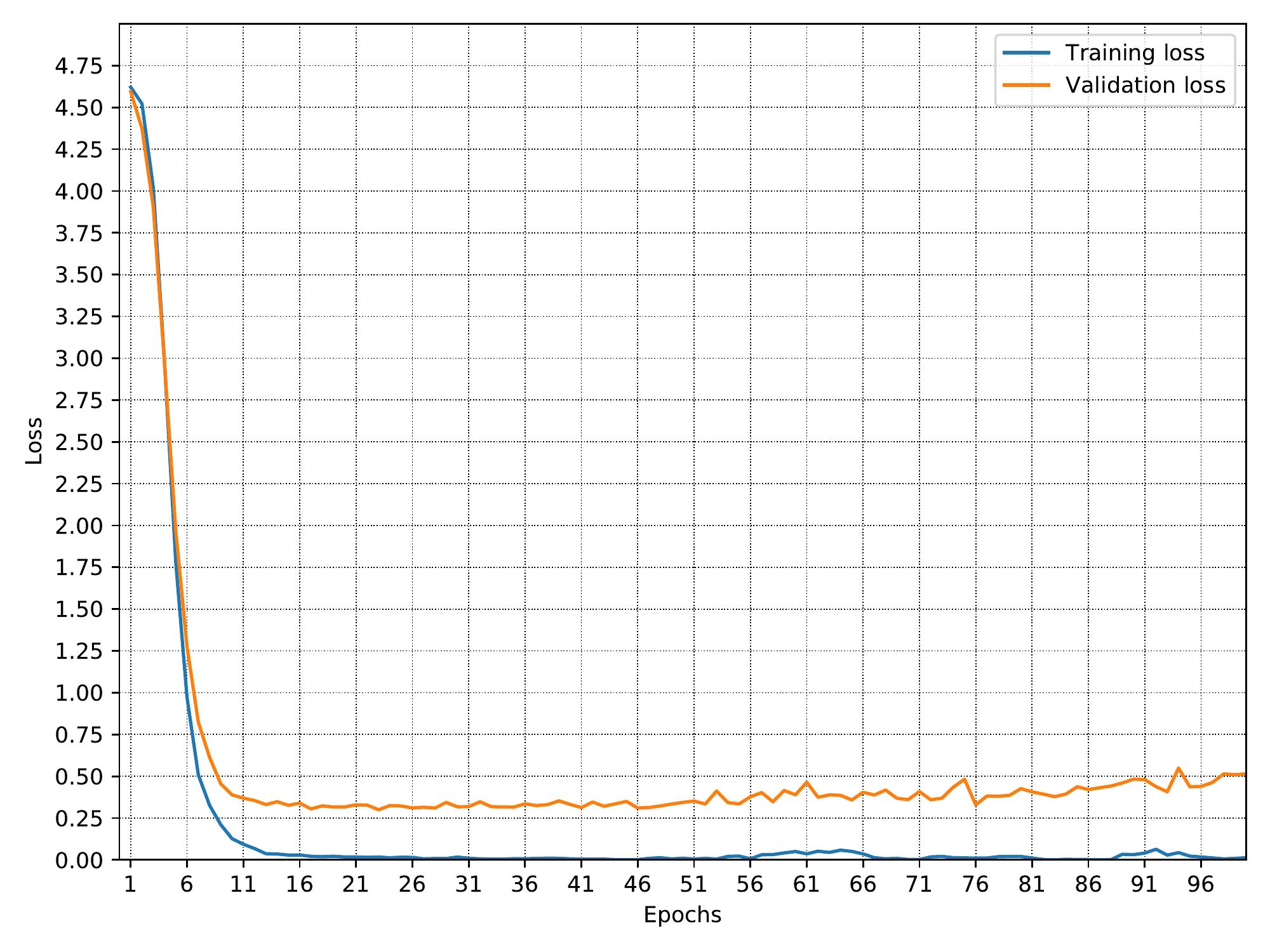}
\caption{Loss variation during learning  time (100 Epochs)}
\label{fig900}
\end{minipage}
\end{figure}

\begin{figure}[htbp]
\centering
\begin{minipage}{9cm}
\includegraphics[width=9cm,height=6.75cm]{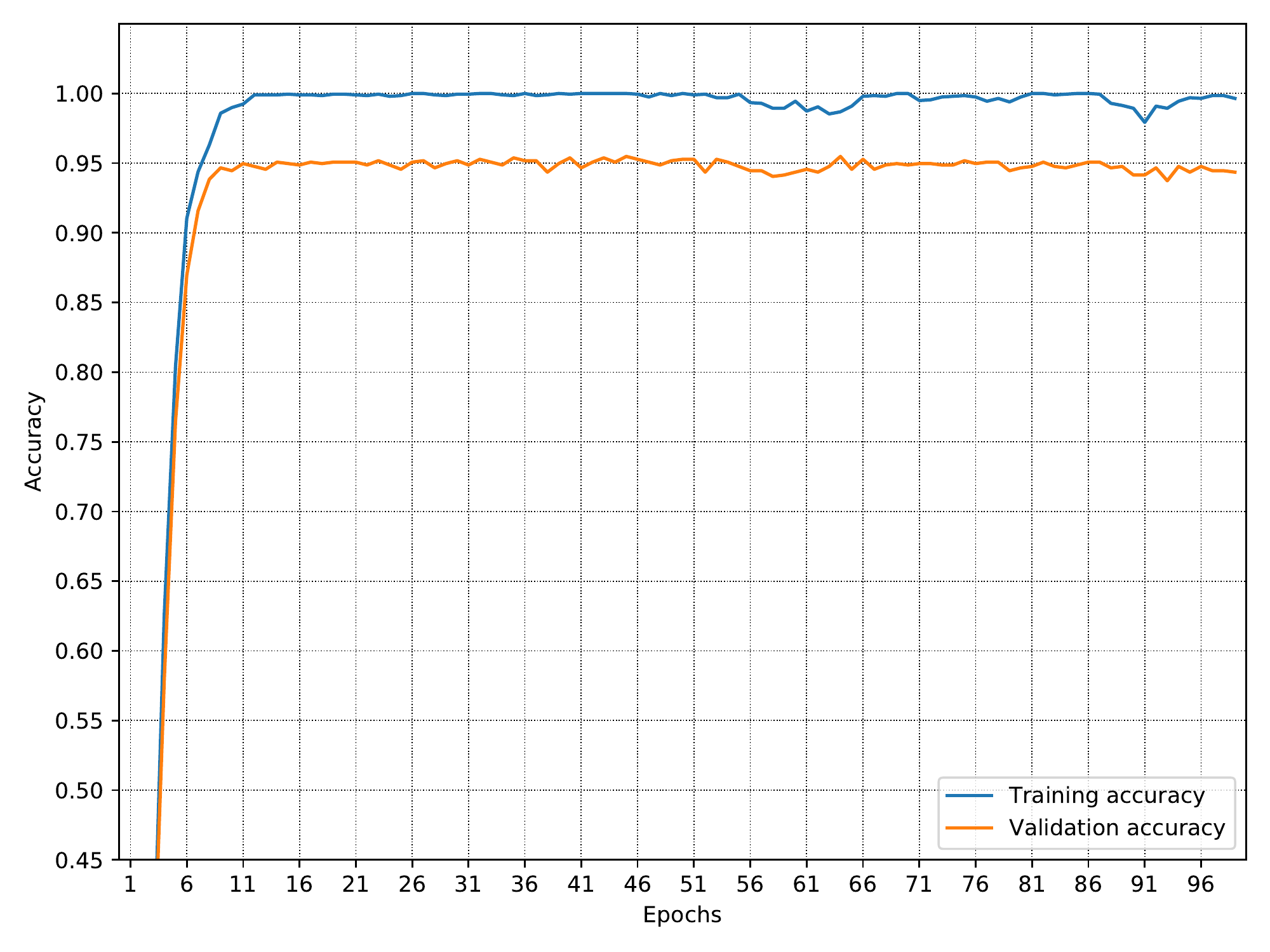}
\caption{Accuracy variation during learning time (100 Epochs)}
\label{fig7}
\end{minipage}
\end{figure}

\begin{figure}[htbp]
\centering
\begin{minipage}{9cm}
\includegraphics[width=9cm,height=7.0cm]{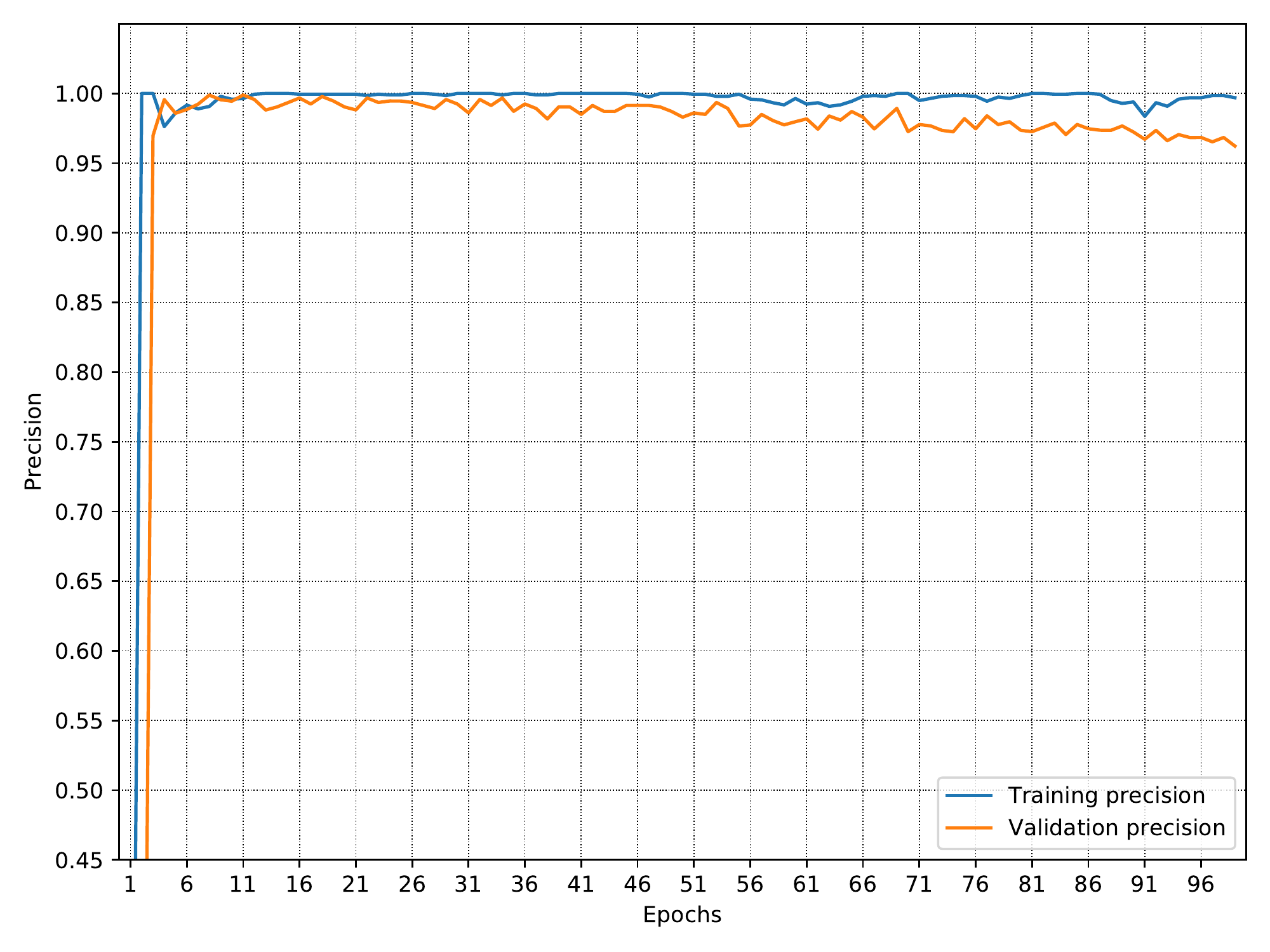}

\caption{Precision variation during learning time (100 Epochs)}
\label{fig8}
\end{minipage}
\end{figure}

\begin{figure}[htbp]
\centering
\begin{minipage}{9.25cm}
\includegraphics[width=9.25cm,height=7.25cm]{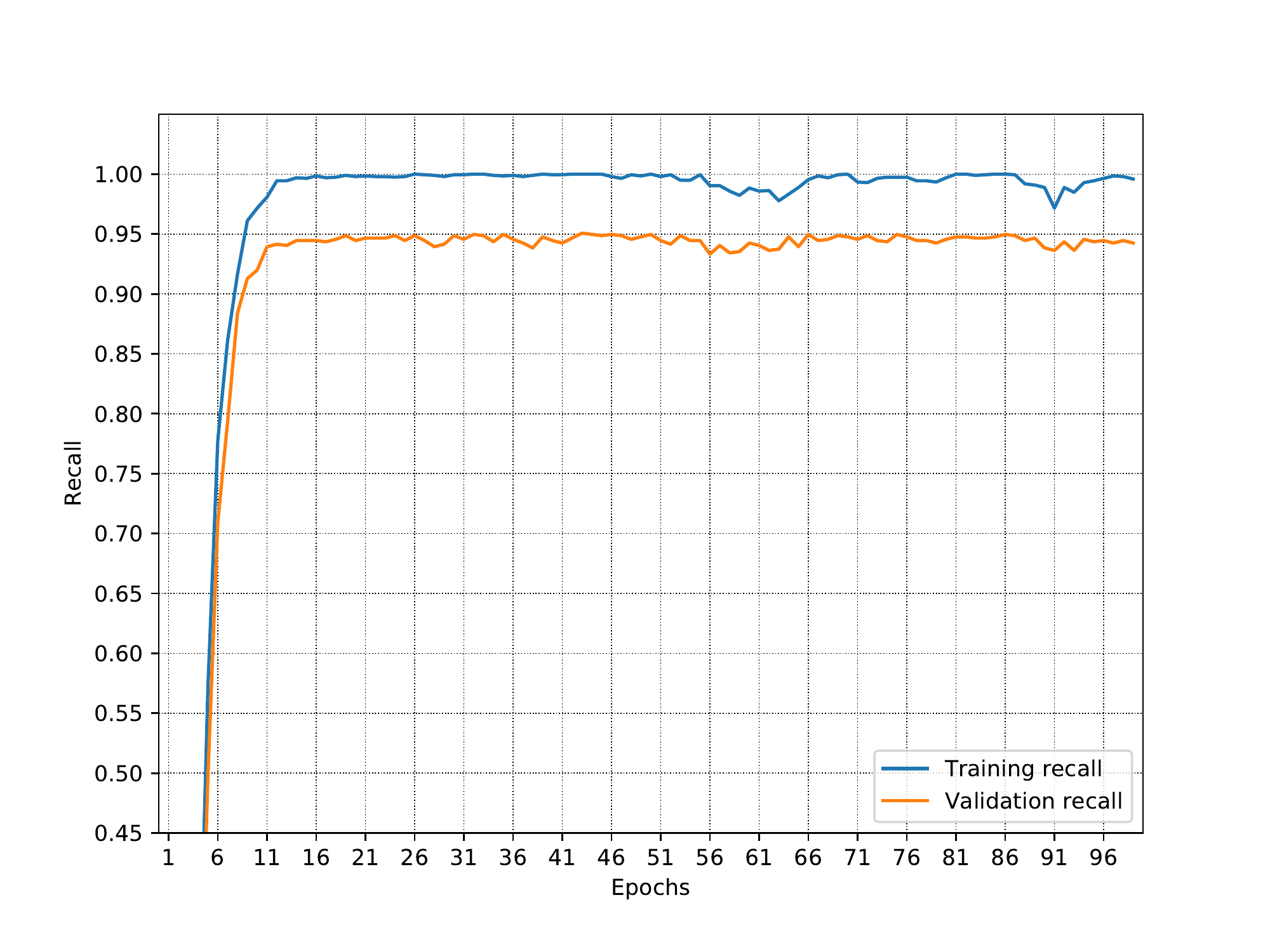}
\caption{Recall variation during learning time (100 Epochs)}
\label{fig80}
\end{minipage}
\end{figure}

\begin{figure}[htbp]
\centering
\begin{minipage}{10.5cm}
\includegraphics[width=10.5cm,height=7.5cm]{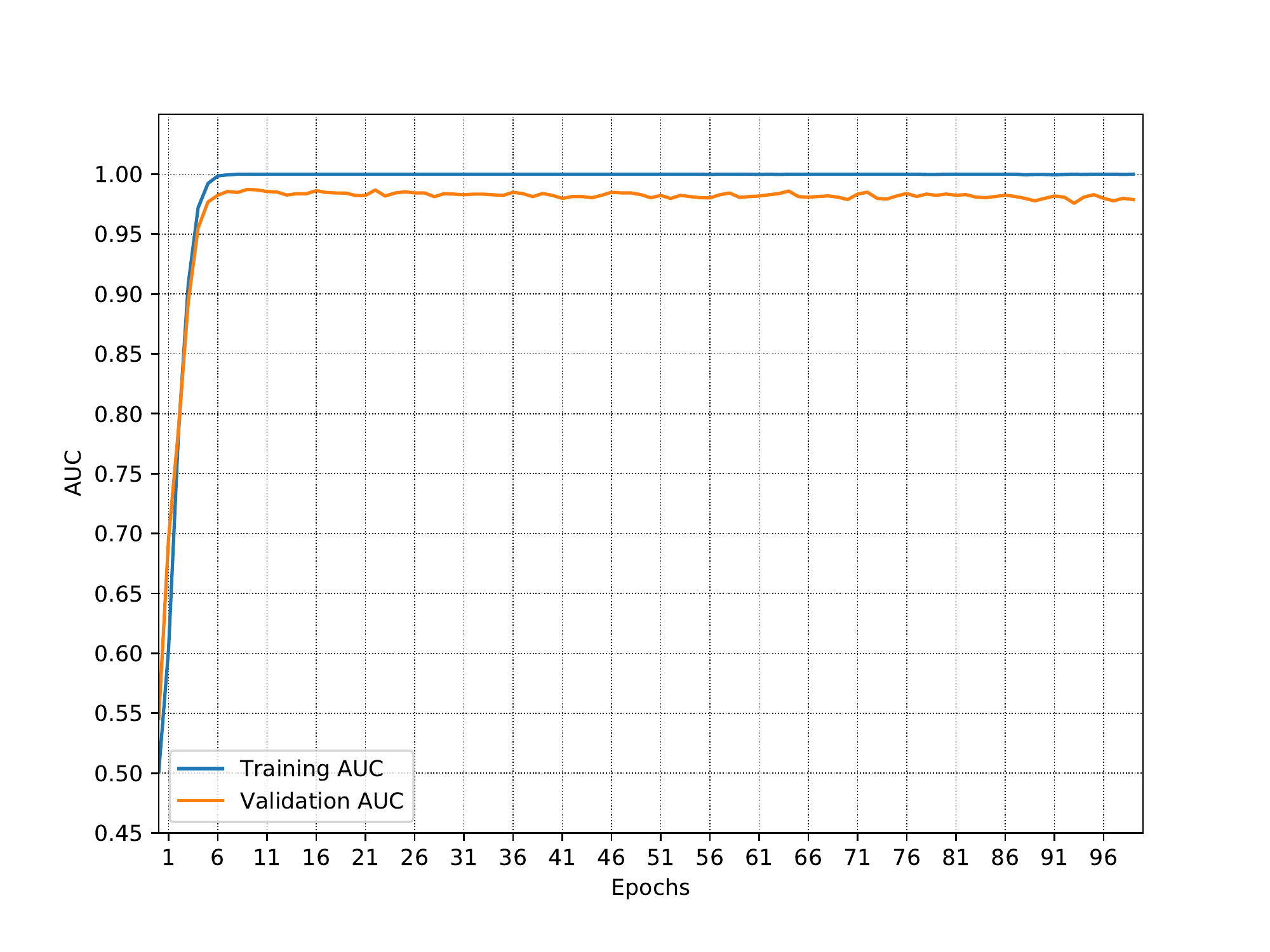}

\caption{Area under the curve (AUC) variation during\\ learning time (100 Epochs)}
\label{fig90}
\end{minipage}
\end{figure}

\begin{center}
\begin{equation} \text{Recall} =\displaystyle \frac{\text{TP}}{\text{TP + FN}} \end{equation} 
\end{center}

where \text{TP, TN, FP, FN} represent the true positives, true negatives,  false positives, and false negatives, respectively.  Accuracy measures the portion of people who are identified correctly by the model through their fingerprints. Precision is a useful metric when the cost of a false positive is high (i.e. in a forensic investigation, this means someone who is wrongly identified as guilty by the model). Recall, in contrast, is useful when a false negative matters (i.e. a criminal not being identified by the model). We also use the Area Under the Curve (AUC) metric to make sure that our model does not randomly predict its outputs. For all of these four metrics, the closer the metric is to 1, the better the model is, for the targeted property. Fig \ref{fig900} illustrates the variation of the loss function (we use the categorical cross-entropy loss function) during learning time. Fig \ref{fig7}, Fig \ref{fig8}, Fig \ref{fig80}, and Fig \ref{fig90}  illustrate the variation of the accuracy (we use the categorical accuracy), precision, recall, and AUC during learning time, respectively.

We notice that the model reaches its optimums at epoch 27, with the values presented in Table     \ref{tab:tablePerfor} obtained on the validation dataset, after reaching 100\% for the four measurements and the minimum value for the loss function over the training dataset.

\renewcommand{\arraystretch}{2}

\begin{table}[h!]
  \begin{center}
    \caption{Model Performance}
    \label{tab:tablePerfor}
    \begin{tabular}{|c|c|c|c|c|} 
      \hline
      \textbf{Model}  & \textbf{Accuracy(\%)} & \textbf{Precision(\%)} & \textbf{Recall(\%)}& \textbf{AUC(\%)} \\
      \hline
      \textbf{ConvLSTM}  & 95.07 & 99.35 & 94.87& 98.83\\
      \hline
      \textbf{LSTM (only)}  & 91.01 & 95.14 & 90.77& 94.83\\
      \hline
    \end{tabular}
  \end{center}
\end{table}

As can be seen in Table \ref{tab:tablePerfor}, our Convolutional LSTM model presents very satisfactory results for all selected metrics in the recognition of subjects with damaged fingerprints, exceeding  95\% accuracy, 99\% precision, and approaches 95\% recall and 99\% AUC. We can also see the clear contribution of Convolutional LSTM compared to conventional LSTM with all metrics. In fact, the convolution layers were able to catch the most important features from the input images. These features were provided to LSTM as sequences, and the latter coped well with them as expected. This synergy between the two layers yielded the good performance of the whole model. Comparative models have proven their performance in similar fields such as scene text recognition \cite{DBLPWangHJHBLL20}, dynamic gesture recognition \cite{DBLPPengTLYL20}, hyperspectral image classification \cite{DBLPHuLPLTD20}, emotion recognition \cite{DBLPHuangLTLY18}, event surveillance  \cite{DBLPZhouZZ17}, traffic forecasting \cite{DBLPLiangWNL20},  air quality prediction \cite{DBLPGuoGCWL19}, facial expression recognition\cite{IEEE8946025}, etc. These good results encourage us to explore the recognition, by comparable networks, of other biometric forms such as hand or earlobe morphology, retinal and iris physiognomy \cite{DBLPLanggartner20}, voice \cite{DBLPGiorgiBEMR19, DBLPKusmierczykSZS19}, and signatures \cite{DBLPChenC2011111}, especially those that have been forged or falsified or altered \cite{DBLPDalilaBSA20}. These biometric patterns play as crucial a role as fingerprints in forensic investigative procedures and in establishing proof of innocence or indictment. The same important role can be played in the identification of victims in disasters and natural catastrophes. That said, as our model has been trained and validated on datasets of hundreds of subjects, when reality involves millions of subjects, it requires more investigation on larger real-world data. This will certainly entail many complexities and challenges that will be the focus of future work.

\section{Conclusion} \label{sec8}

Our research results in a well performing recurrent neural network model that can be used to examine and interpret evidence in criminal proceedings involving fingerprints, deliberately damaged by seasoned criminals. Our model can help prove the guilt or innocence of suspects. It can also help resolve a wide range of legal issues related to the identification, analysis and proof assessment. We have also mapped out future paths that involve similar models as well as other comparable biometric forms that can be of great value in an investigation process.

\section*{Acknowledgment}

This research was funded by the Natural Sciences and Engineering Research Council of Canada (NSERC).

\bibliography{biblio}

\bibliographystyle{IEEEtran}

\end{document}